\documentclass[conference]{IEEEtran}
\usepackage{csquotes}
\usepackage{cite}
\usepackage{amsmath,amssymb,amsfonts}
\usepackage{algorithmic}
\usepackage{graphicx}
\usepackage{textcomp}
\usepackage{subfig}
\usepackage{url}
\usepackage{todonotes}
\usepackage{xcolor}
\def\BibTeX{{\rm B\kern-.05em{\sc i\kern-.025em b}\kern-.08em
		T\kern-.1667em\lower.7ex\hbox{E}\kern-.125emX}}
\begin{document}
\title{Procedural Content Generation: Better Benchmarks for Transfer Reinforcement Learning}

\author{\IEEEauthorblockN{Matthias Müller-Brockhausen, Mike Preuss, Aske Plaat}
	\IEEEauthorblockA{\textit{Leiden Institute of Advanced Computer Science} \\
		\textit{Leiden University}\\
		The Netherlands \\
		m.f.t.muller-brockhausen@liacs.leidenuniv.nl}
}

\maketitle
\begin{abstract}
The idea of transfer in reinforcement learning (TRL) is intriguing: being able to transfer knowledge from one problem to another problem without learning everything from scratch. This promises quicker learning and learning more complex methods. 
To gain an insight into the field and to detect emerging trends, we performed a database search. We note a surprisingly late adoption of deep learning that starts in 2018.
The introduction of deep learning has not yet solved the greatest challenge of TRL: generalization. Transfer between different domains works well when domains have strong similarities (e.g. MountainCar to Cartpole), and most TRL publications focus on different tasks within the same domain that have few differences. Most TRL applications we encountered compare their improvements against self-defined baselines, and the field is still missing unified benchmarks. We consider this to be a disappointing situation. For the future, we note that: (1) A clear measure of task similarity is needed. (2) Generalization needs to improve. Promising approaches merge deep learning with planning via MCTS or introduce memory through LSTMs. (3) The lack of benchmarking tools will be remedied to enable meaningful comparison and measure progress. Already Alchemy and Meta-World are emerging as interesting benchmark suites. We note that another development, the increase in procedural content generation (PCG), can improve both benchmarking and generalization in TRL. 
\end{abstract}


\section{Introduction}
\begin{figure}
    \centering
	\includegraphics[width=0.5\textwidth]{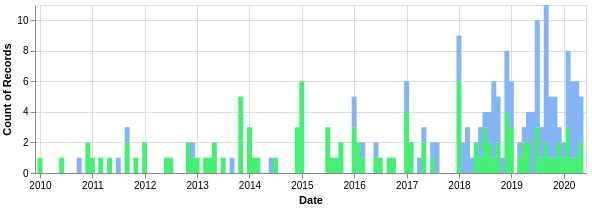}
	\caption{Number of published papers per month. Light blue indicates entries that make use of deep neural networks, and green ones do not. The years 1985 to 2009 are cut off for readability, but the full graph is available at~\cite{interactivewebsite}.}
	\label{fig:pub}
\end{figure}

The need for Transfer in Reinforcement Learning (TRL) is growing through increased usage of deep reinforcement learning (RL), as shown in Figure \ref{fig:pub}. Deep networks are expensive to train~\cite{patterson2021carbon}, so cutting down the learning time by re-using previously gained knowledge is desirable.
Computer games are currently often used as benchmarks or challenging test systems, and here it is especially the case that changes of different amplitude (e.g. patches) happen regularly. Recent AI successes on well-known complex games such as Dota2~\cite{dota2} and StarCraft~II\cite{vinyals2019grandmaster} show that constant retraining is necessary as the underlying systems evolve on the same time scale as the trained AI systems.

Although the need for knowledge transfer in a reliable manner is clear, most experiments show limited generalization, and progress in TRL is limited. Our main goal in this paper is to detect why this is the case and how it can be cured.
It turns out that a major problem lies in the sparsity of suitable benchmarks, and we see the use of Procedural Content Generation (PCG) as a recommended solution to this problem. 

This paper has the following contributions. 
\begin{enumerate}
    \item We categorize the literature of Transfer in Reinforcement Learning (TRL), finding many different approaches and applications
    \item The absence of clear benchmarks and a clear research agenda is noted
    \item We provide a research agenda in which we stress the need for a clear measure of success, clear benchmarks, and suggest that Procedural Content Generation is ideally suited to provide such benchmarks for transfer reinforcement learning
\end{enumerate} 
Section \ref{sec:relatedwork} introduces related work (a meta-survey). To gain an unbiased insight into TRL, we scrape through a dataset (Section \ref{sec:method}). We explain the experimental parameters and decisions involved in TRL and identify trends in their usage (Section \ref{sec:analysis}).
After categorizing a large number of transfer experiments, we report the generalization capabilities of TRL (Section \ref{sec:generalization}). 
We draft a research agenda for TRL (Section \ref{sec:outlook}) that lists current limitations and identifies directions that the field is likely headed to.

\section{Related Work}
\label{sec:relatedwork}
\begin{figure*}[h]
	\centering
	\includegraphics[width=0.75\textwidth]{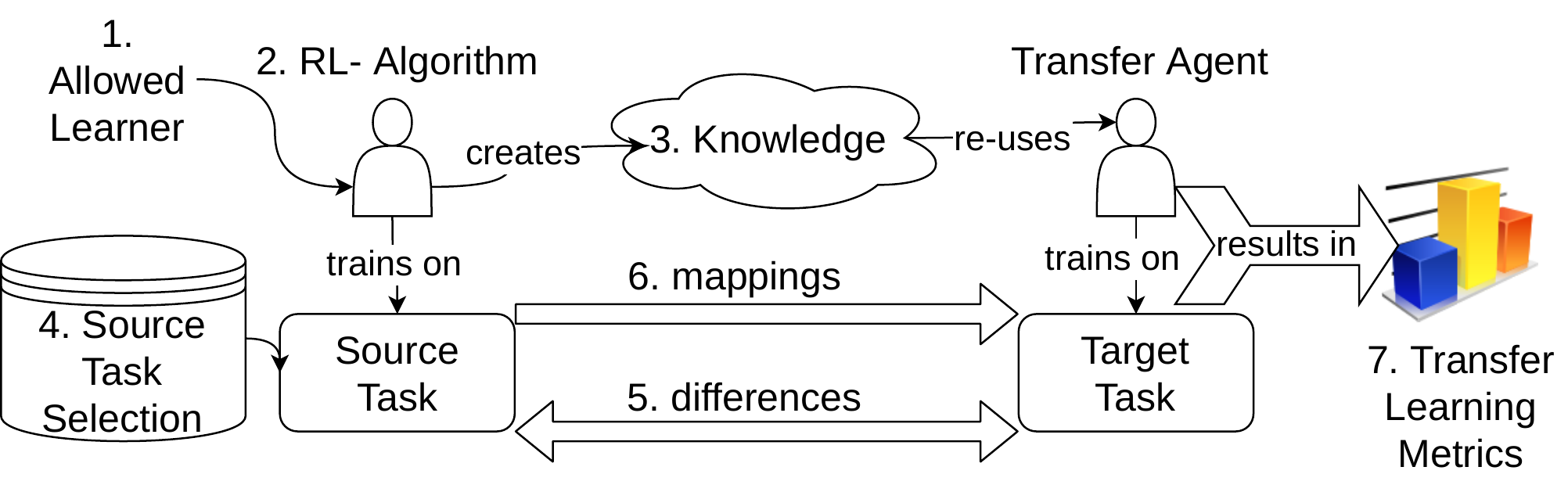}
	\caption{A visualization of the transfer learning process. The chosen allowed learner (1) / algorithm (2) combination generates knowledge (3) while training on a selected source task (4). The source and target task have to differ (5), and depending on how large these differences are, task mappings (6) might be required. By using the previously gained knowledge, the transfer agent then trains on the new task. The gathered data from source and target training can be distilled into the transfer learning metrics (7).}
	\label{fig:overview}
\end{figure*}

Some summaries on individual aspects of Transfer in Reinforcement Learning (TRL) exist, and we will introduce them in chronological order. The first one was written by Bone in 2008~\cite{bone-survey}, still pre-deep learning.
A year later, Taylor \& Stone followed~\cite{taylor-survey} with a comprehensive survey. Its authors are the two most recurring names in the field in our data set. Two years later, they published a second survey with a focus on inter-task transfer~\cite{taylor-intertask}. In 2012, Lazaric formulated a framework~\cite{lazaric-survey} that enables the categorization of TRL experiments similar to~\cite{taylor-survey}. Seven years later, in 2019, Da Silva \& Costa published a survey focused on multiagent RL~\cite{silva-survey}.
In 2020, multiple surveys appeared. One on curriculum learning in RL~\cite{cur-survey}, one about multi-task transfer~\cite{multitask-survey}, and one about transfer in deep reinforcement learning~\cite{zhu-survey}. Multiple surveys in one year might seem peculiar, but a look at Figure \ref{fig:pub} shows a large increase in publications starting in 2018.

These surveys formulate frameworks to encompass the different TRL literature they encounter~\cite{taylor-survey,lazaric-survey}, or analyze specific sub-fields of TRL~\cite{taylor-intertask, silva-survey, cur-survey, multitask-survey, zhu-survey}. We aim for a comprehensive overview of all aspects of TRL. We perform scraping through the Microsoft Academic Graph~\cite{mag}, which contains over 209 Million papers, and include papers based on keywords in title and abstract. This yields some interesting statistics. For example, $\approx74.8$\% of the 270 relevant papers in our data set have not been included in any previously mentioned survey~\cite{zhu-survey,lazaric-survey,taylor-intertask,bone-survey,silva-survey,taylor-survey,cur-survey,multitask-survey}.
Moreover, it enables visualizations such as a publication timeline (Figure \ref{fig:pub}), a social network graph (Figure \ref{fig:community}), and the creation of an interactive web tool~\cite{interactivewebsite} that facilitates the re-use of the data.
Automatic quantitative analysis is to be seen as an addition to produce insightful figures and tools. It also served as tool to gain insight into the state of the field. But to form a true vision for our research agenda (Section \ref{sec:outlook}) we still relied on manual research.

\section{Method}
\label{sec:method}
We search through a snapshot~\cite{mag-supply} of the Microsoft Academic Graph (MAG)~\cite{mag} for entries that contain the three words "transfer," "reinforcement" and "learning" in either their abstract or title.
We create a spreadsheet using the terminology introduced by~\cite{taylor-survey}:
transfer dimension, allowed task differences, source task selection, task mappings, transferred knowledge, allowed learners, transfer metrics. We also record which RL algorithm was used (e.g., Q, DQN, PPO, DDPG), whether the paper publishes additional resources such as source code, and we check if the links work~\cite{zittrain2014perma}. 
All data and resources, including the spreadsheet, and an interactive data viewer, are openly available at~\cite{interactivewebsite}.

%\section{Background Information}
%\label{sec:background}

Figure \ref{fig:overview} provides a visualization of the process of a TRL experiment. For terminology, we stat as close as possible to~\cite{taylor-survey}. 
The RL-algorithm (2) provides more details about the learning algorithm that is used (1) and how transferred knowledge is re-used. We briefly review which information was gathered from the papers. The content in parentheses behind keywords refers to the transfer process step in Figure \ref{fig:overview} if numerical, or otherwise the abbreviation used in the spreadsheet. By explaining the Figure, we also follow the workflow of setting up a typical TRL experiment, and we describe essential choices of the authors.

The allowed learner (1) places restrictions on how transfer is approached and influences experimental parameters such as the pool of available reinforcement learning methods (2). The options are temporal difference learning (TD), model-based learning (MB), relational learning (RRL), hierarchical learning (H), batch learning (Batch), Bayesian methods (bayes), and case-based reasoning (CBR). Because many different algorithms exist per allowed learner method, we also note which RL-algorithm was used (2).
The next step is to determine the type of knowledge (3) to transfer. The easiest methods re-use what was learned, e.g., the action-value function (Q), policy ($\pi$), task model (model), found prior distributions (pri), or experience instances (I).

However, higher-level knowledge can also be used in a variety of ways. One can extract partial policies ($\pi_p$) or options (options) for guidance.
Rules or advice from experts (ra), be it human demonstration or successfully trained agents, can guide the training process. This advice can manifest itself in a shaped reward (R). Some algorithms can identify and learn important features (fea), autonomously find sub-task definitions (sub), or build a proto-value function (pvf).
Other methods to transfer knowledge are a Variational auto-encoder (VAE)~\cite{yang2020single} or policy distillation~\cite{multipolar}.

To gather the re-usable knowledge,\footnote{Simplified for the Figure. Knowledge used to train the transfer agent can stem from anywhere, like human made demonstrations or rules.} an agent needs to train on a selected source task (4). Either all previously seen tasks are used (all), one is selected by the author (h), a library of tasks to choose from is defined by the author (lib), or the agent has to modify the source task to gain the required knowledge autonomously (mod).
There are two important factors on transfer between the source and target task. The first are differences (5) an agent has to handle between tasks. These can be small, such as an alternating start ($s_i$) or end ($s_f$) position, another level layout, or a different number of encountered objects ($\#$). Changes can also affect the number of involved objects (\#), transition function (t), state variables (v), the action set (a), or reward function (r).
Secondly, mapping between tasks (6) could be required. The agent can get no mapping (N/A) or learn it from experience (exp). The mapping can also be provided as higher-level knowledge (T), created by humans (sup), derived from action mapping ($M_a$), or a grouping of state variables ($sv_g$).

The transfer experiment results in metrics (7) that indicate success. One can measure an improvement in the Time to Threshold (tt), so how many fewer steps did the agent have to train to reach a previously specified reward threshold. If the transfer agent training starts at a higher reward than an agent that started from scratch, one has measured a Jumpstart (j). The transfer agent can also achieve a higher total reward (tr), and the difference between transfer agent and training from scratch is called transfer ratio\footnote{We omit the transfer ratio as performance improvement measurement in our data and in Figure \ref{fig:success}, as it is an extension of the total reward~\cite{ammar2013automated}.}. Lastly, Asymptotic Performance (ap) indicates whether the final learned performance has improved.

\section{Analysis}
\label{sec:analysis}
To get an overview of the dataset, the social network structure of the citation data is visualized (section \ref{sec:networks}), and insights from the categorization are presented (section \ref{sec:categories}). The main goal of transfer learning is generalization. We analyzed the papers for the strength of generalization that has been achieved in TRL (section \ref{sec:generalization}).

Many more insights have been found, and we refer the reader to discover the full data on an interactive website~\cite{interactivewebsite}.

\subsection{Social Network Structure}
\label{sec:networks}
\begin{figure}
	\centering
	\includegraphics[width=0.48\textwidth]{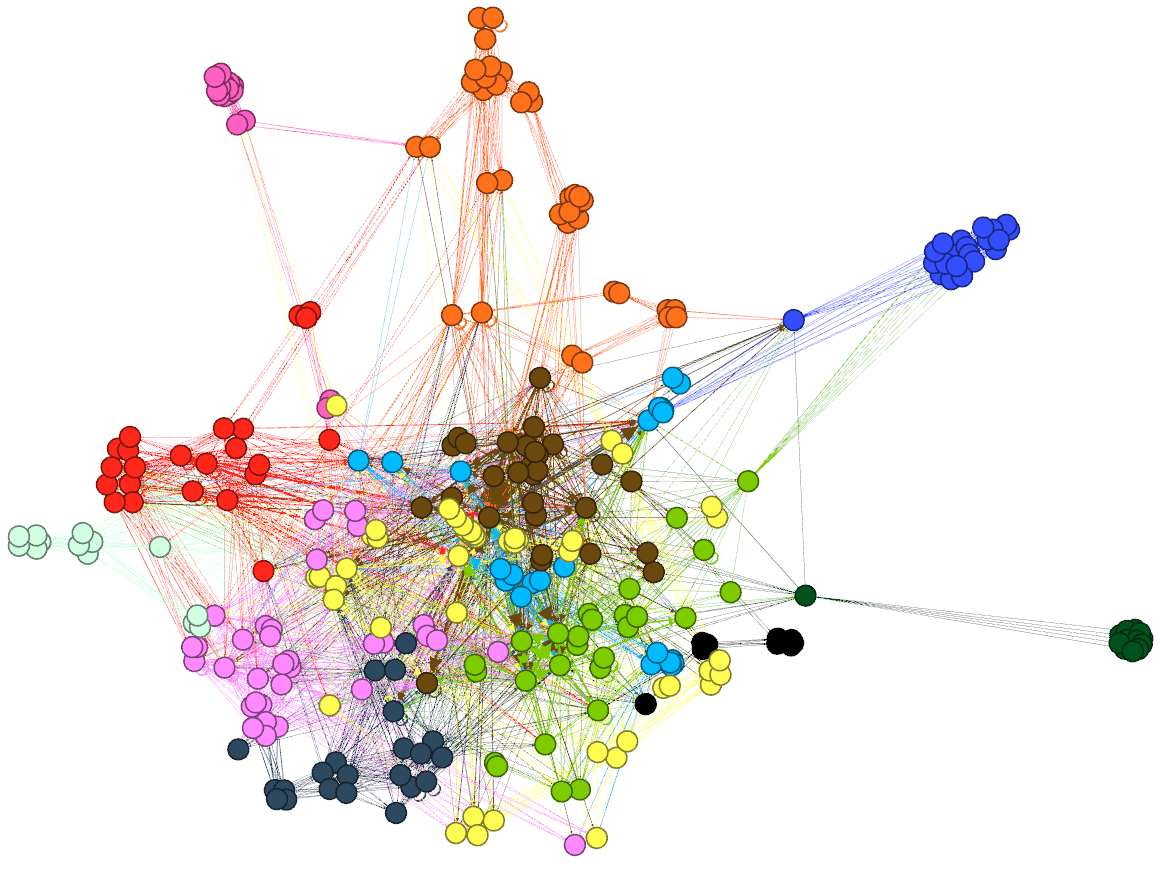}
	\caption{Directed Social Network Graph of Authors citing each other. The red group contains mostly case-based reasoning learners, purple contains hierarchical algorithms and Bayesian RL, dark green is the Sim2Real community, black papers all applied Q-learning, dark blue contains reward shaping, orange reports mostly on applications (such as energy consumption in buildings). Layout determined by Force Atlas~\cite{jacomy2014forceatlas2}.}
	\label{fig:community}
\end{figure}

Figure \ref{fig:community} shows a visualization of citations between authors. Colors indicate different communities, as identified by the community detection algorithm~\cite{gephi-community}. 
Note that due to noise and missing values in the base data, 63 ($\approx23\%$) of the entries are missing reference data and are therefore not present in Figure~\ref{fig:community}.

Eight connected components have been identified. Only one of these contains most entries. The other seven are independent. The independent components are not included in Figure \ref{fig:community}.
The largest component consists of nine individual communities.
For each community, we attempt to identify common aspects and succeeded for four of them.
Two communities focus on the different types of allowed learners. Red at the far left contains mostly case-based reasoning (CBR), and purple at the bottom left hierarchical (H) and Bayesian RL (bayes).
The dark green community at the far right contains mostly Simulation to Reality (Sim2Real) experiments. From here on out, the similarities are already declining. As for the black group, the only link we could find is that they all applied Q learning at some point.

In addition to hierarchical learners, purple also contains numerous 2D navigation experiments, but not exclusively. Dark blue at the bottom contains many reward shaping experiments that sample from a (human) demonstration or world model prediction.

For the remaining five communities, no real common aspects could be found, except that $\approx44\%$ in the orange community focus on real-world engineering problems such as optimizing power usage for buildings~\cite{mocanu2016unsupervised}, air conditioners~\cite{hvac}, or collision avoidance for autonomous vehicles~\cite{miriti2014integrating}. % 16 of 36 = 44

\subsection{Category Data}
\label{sec:categories}
Of the 270 entries, 202 ($\approx74.8\%$) have not been in previous surveys of the field~\cite{zhu-survey,lazaric-survey,taylor-intertask,bone-survey,silva-survey,taylor-survey,cur-survey,multitask-survey}. In the following, parentheses will indicate the number of papers in the dataset relating to a specific message, for example $10\%$ (27) entries have not been approved through peer-review (23 arxiv, 4 rejected on openreview). Text in brackets refers to variable abbreviation presented in Section \ref{sec:method}. %We want to congratulate and thank the 14 Ph.D.- and 5 Master-Theses for advancing the field.
Furthermore, one entry may contain multiple tags per category.
We follow the order of the steps of the transfer learning process of Figure \ref{fig:overview}) in presenting the data.

First, the allowed learner is chosen. The majority of papers uses regular temporal difference methods (241) [TD], followed by hierarchical learning (46) [H], model-based learning (31) [model], Bayesian learning (20) [Bayes], batch learning (13), relational learning (11) [RRL], policy search learning (10) [PS], case-based reasoning (9) [CBR], and one linear programming entry. TRL transfers knowledge between different tasks, and it is no surprise that hierarchical learning is the second most applied learner type for TRL because the multiple tasks might be hierarchically related. Moreover, singular large tasks can be decomposed into multiple (hierarchically ordered) sub-tasks for transfer~\cite{taylor-survey}.

The allowed learner narrows the pool of RL algorithms that can be chosen. The most popular algorithms are tabular Q-Learning (124), DQN (36), SARSA (28), DDPG (11), PPO (10), FQI (10), DDQN (8), A3C (7), LSPI (7), and Policy Gradient (6).
The high occurrence of tabular Q-Learning could give the impression that deep learning is not prevalent in TRL yet, but $\approx36\%$ (99) already use deep neural networks. Moreover, Figure \ref{fig:pub} depicts a clear trend towards deep learning that started in 2018. 

Although the number of 3D environments that deep neural networks (DNN) (17) are applied to are almost the same as for tabular algorithms (14), their complexity differs. Tabular algorithms focus mostly on balancing problems, such as Mountain Car 3D (11) or controlling joints in a real-world RoboCup robot~\cite{barrett2010transfer, celiberto2011using}. The 3D environments that DNN’s are applied to are more complex. AirSim requires full free 3D navigation through flying, plus the policy is transferred to a real drone~\cite{yoon2019hierarchical}, and Mujoco combines controlling multiple joints with moving in a 3D World~\cite{xie2018transferring}.

The third step (see Figure) is to decide what knowledge should be extracted or transferred. The most popular methods are also the easiest to transfer, namely just re-using the previously learned action-value function (72) [Q] or policy (64) [$\pi$]. All other methods require more sophistication, such as extracting and re-using relevant features (38) [fea] or guiding training via Advisor / Teacher data (22) [advisor]. Shaping the reward (20) [R] is also an often-used means to transfer knowledge. Less often used are approaches like collecting experience instances (14), building a task model (14) or sub-task definition (8), defining rules (14), finding options (12), or collecting distribution priors (9). There are also new ways to transfer knowledge between tasks. One of these is policy distillation (4). Most entries used it to summarize multiple learned policies into a single one~\cite{yin2017knowledge, multipolar, xiao2018distributed}, but it can also be used for simulation to reality transfer~\cite{traore2019continual}. The distillation process and the advisor method have one element in common: They both re-use the same type of saved knowledge.
Another transfer method that was introduced through deep neural networks is the Variational Auto Encoder (VAE). These networks can help to automatically identify relevant features in the latent space and thus allow for universal control policies~\cite{yang2020single}.
Moreover, Q-functions can be generated algorithmically~\cite{arnekvist2019vpe}. This is related to hyper-networks, where a neural network outputs the weights used in another network~\cite{stanley2009hypercube}.

The fourth step is determining the source task from which knowledge should be transferred. Here most approaches perform hand-selection (137) [h]. 48 entries let the algorithm use all tasks (all), 32 use a library of select tasks (lib), and in two cases, the agent automatically modifies a provided source task (mod).

The fifth and most important question to answer is: What kind of differences in tasks will the learner have to handle? The most frequently occurring differences are in transition dynamics (115) [t] of the environment. Different transition dynamics refer to changes in parameters between tasks that influence how the world changes per timestep. For example, if the agent were to control an airplane, changes in gravity, weight, or friction would count as transition dynamic differences.
The second to fourth place goes to navigation-related tasks, namely changes in the goal- (95) [$s_f$] or start- (91) [$s_i$] position or the level layout (56).
Values that the agent receives are also well suited to accelerate the transition. For that, the received observation (52) [v], the action set (43) [a], the number of objects encoded in the observation (32) [\#], or the reward (29) [r] given to the agent can be adjusted.

Based on the task differences, the sixth step, namely mappings between tasks, must be determined. They are not often necessary as 164 do not use any mapping between tasks. Nevertheless, when they are used, the majority of algorithms try to learn them from experience (40) [exp], or they are given manually (25) [sup]. The less-used methods are action mappings (8) [$M_a$] or groupings of state variables [$sv_g$]. One interesting approach for mappings in image-based domains is the Generative Adversarial Network (GAN)~\cite{gamrian2019transfer}.

\begin{figure}
    \centering
	\includegraphics[width=0.4\textwidth]{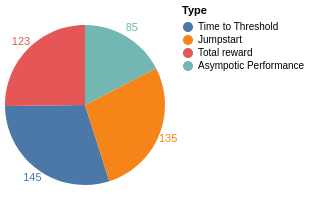}
	\caption{The number of times that papers reported that transfer performance improvements (Section \ref{sec:method}) were measured by category.}
	\label{fig:success}
\end{figure}
The last step in transfer learning, and the most important to get an idea of the experimental success, are the associated metrics (Figure \ref{fig:success}). 146 entries achieved a decrease in the time to reach a threshold (tt), while 135 were able to jump-start (jp) the reward in a new setting. 123 entries achieved a higher total reward (tr) compared to no transfer at the end of training, and 85 entries trained agents that show asymptotic performance (ap) after transfer. Many papers measured multiple metrics of success. 91 achieved two, 38 three, and 27 all four metrics.

We also looked at what kind of problem TRL is mostly applied to. The most often recurring applications are navigation (122), robotics (56), classic control (42), and games (26).
The navigation domain is the most diverse. The majority of experiments inspect 2D (110) instead of 3D (12) worlds. To further simplify 2D worlds, 79 entries use a grid instead of continuous navigation. Most entries formulate their own problem, but there are also some recurring standardized environments such as the Taxi world from Dietterrich~\cite{dietterich2000hierarchical} or the blocks world by Langley~\cite{blocksworldcreator}. Although most 2D grid-level layouts do not cite anyone, there is one recurring citation: the three-room grid-level by Thrun~\cite{schwartz1995finding}. One entry also uses a slightly adjusted version of the three-room grid-level~\cite{arabasadi2014learning}.
The goal of these papers is to test if transfer is possible. We would expect test domains to be different, challenging, and standarized. The state of affair that we encountered is lacking in this respect. While some problems repeat, there is no unified benchmark, and the few existing benchmarks are not dynamic in the sense that they could adapt their difficulty or similarity (Section \ref{sec:outlook}).
 Given the large number of entries on the topic, we were surprised that there is no real benchmark to assess the planning capabilities of an RL algorithm in the navigation domain.
The ProcGen environment is such a benchmark for maze navigation~\cite{cobbe2019procgen}. However, ProcGen is an environment with procedurally generated levels, and most entries here use one (or more) static levels. And~\cite{justesen2018illuminating} has found that while (deep) RL can learn to generalize to generated levels within the same distribution, it can not handle arbitrary level layouts.

We encountered 27 game-related entries, of which 11 focus on Atari and only 4 on board games.
Few complex games are approached with TRL, like Unreal Tournament~\cite{hou2017evolutionary}, StarCraft~\cite{shao2018starcraft}, or GVGAI~\cite{narasimhan2018grounding}. Simpler games include Sonic 3~\cite{hamilton2020sonic} or Pinball~\cite{yang2020efficient}.

The growing use of neural networks comes with a drawback, namely the reproducibility of results. The ICLR Reproducibility Challenge~\cite{reprochallenge} from 2018 underlines the problem, as less than 33\% of papers are rated as properly reproducible~\cite{reprotalk}. The most straightforward way to make a code-based science experiment reproducible is by publishing the source code. Only 15 entries did this, but at least 148 contain pseudo-code.
Moreover, 31 entries link additional resources. 10 of these are websites, but 8 of them are not available anymore. The two websites that are still reachable summarize different short videos of robotics tasks on one page. The remaining 21 links are videos.

Another important aspect related to reproducibility is the software libraries used in the experiment code. Even when closed source, some entries do share details about used libraries. As machine learning backend, TensorFlow (20) and 7 PyTorch (7) are popular. Only 4 of the TensorFlow uses are from DeepMind, so the library's popularity seems community-driven. Regarding RL environments, 24 use OpenAI's Gym~\cite{openaigym}, 5 the proprietary physics playground Mujoco~\cite{todorov2012mujoco}, 4 the unreal engine based 3D navigation simulator AirSim~\cite{airsim2017fsr}, and 2 the continuous control benchmark RL-Lab~\cite{duan2016benchmarking}.

\subsection{Generalization Capabilities}
\label{sec:generalization}
Achieving generalization by transferring knowledge from one task to another remains challenging. The more different tasks are, the harder generalization becomes. One algorithm, such as AlphaZero, can learn to play at world champion level in three different board games~\cite{alphazeromultigames}. The limitation is that one network has to be trained again for each game. Unlike humans, the AlphaZero AI can not yet generalize and transfer knowledge from one domain to another similar domain, even when the internal network architecture is identical.
The field of TRL revolves around finding algorithms to enable this transfer between different tasks in the same domain.
Transfer between differing domains works better the more similar they are, and when only the transition dynamics change.

For example, transferring a Q-learner from MountainCar to Cartpole works well~\cite{ammar2013automated}. Also, transfer from CartPole to Bicycle works well~\cite{joshi2018cross} or a three-linked CartPole to the Quadrotor~\cite{ammar2015unsupervised} control tasks.
Another popular transfer domain is RoboCup, with 24 entries. Many of the experiments focus on increasing the number of players involved from 3v2 to 4v3~\cite{soni2006using} or 3v2 to 6v5~\cite{partalas2009transferring} in KeepAway. Others explore multi-task experiments such as MoveDownfield to BreakAway~\cite{torrey2006skill}. 

Although the RoboCup challenge could be seen as 2D navigation, it does not involve the same amount of planning as is required to navigate through a 2D maze, whether it is grid-based or continuous. For 2D maze navigation, which is intuitive for most humans, reinforcement learning needs special help. For example, repositioning doors in a level whose layout has not changed requires advisors~\cite{plisnier2019transfer}. Picking up a sequence of keys and then moving through doors can be solved by adding options~\cite{konidaris2007building}. For regular search algorithms such as A* these task changes would be easy to solve. However, for large, complicated 2D and 3D navigation domains, try to incorporate planning into RL. There already two great examples of this. First of all, Go-Explore~\cite{goexplore}. Many of the Atari games that~\cite{goexplore} tackles can be viewed as multi-level 2D navigation tasks, such as Montezuma's Revenge, Berzerk, and Private Eye. Go-Explore effectively combines planning with regular reinforcement learning and is good at these navigation and planning Atari games. Other promising approaches exist such as MuZero~\cite{schrittwieser2020mastering} or MCTSnet~\cite{guez2018learning}.

Achieving reliable transfer of knowledge gained from perfect simulations to the real-world is another unsolved transfer problem. We found 23 entries in this sub-field (Sim2Real). Many of these focus on moving joints (3), which could be compared to classic control tasks. It can also be extended to multiple joints that form a robotic arm (12) to interact with objects.
To narrow the gap between simulation and reality, in many papers, noise is applied for regularization or smoothing, e.g., Gaussian Noise~\cite{higgins2017darla}, uniform random noise~\cite{anwar2020autonomous}, the Simulation Optimization Bias (SOB)~\cite{muratore2019assessing}). There are also efforts to make the noise obsolete~\cite{kaspar2020sim2real}.

The findings underline that TRL only works well when some similarity between source and target tasks can be found. In this sense, generalization is in the eye of the beholder, and there is a long way to go. Nevertheless, one paper's generalization capabilities are impressive: By encoding the Video and Audio output of an Atari game into a multi-modal latent space, a policy was trained on video-only that can transfer its performance to audio-only input~\cite{dark}.

\section{Research Agenda}
\label{sec:outlook}
In this literature review, we have categorized around 300 papers on transfer reinforcement learning. We have seen many different approaches trying to transfer knowledge between many different applications. The measure of success is generalization: how well knowledge can be transferred between different applications. We note (1) in supervised learning, transfer has achieved more success~\cite{weiss2016survey} than in reinforcement learning. TRL is still a young field. The first deep learning TRL papers appeared around 2010, but only in 2018 did the field really start adopting it. Deep learning methods can be expected to continue to yield good performance. Transfer reinforcement learning should continue to focus on transfer of network parameters. (2) The large diversity in applications and methods makes progress comparisons difficult. Also, we noted a lack of dynamic benchmarks (Section \ref{sec:categories}). (3) Generalization is hard, except when applications are clearly related. 

These observations bring us to the following research agenda. 
\begin{enumerate}
\item A clear measure of transfer capabilities is needed in transfer reinforcement learning~\cite{carroll2005task, ammar2014automated}. This implies a universal measure of similarities between tasks/domains.
\item The combination of planning and learning can be expected to improve (as already shown by Go-Explore~\cite{goexplore} and Mu-Zero~\cite{schrittwieser2020mastering}). Transfer reinforcement learning should focus on general planning methods~\cite{guez2018learning}.
\item Benchmarks are needed that are standardized, challenging, and dynamic (Section \ref{sec:categories}). Procedural Content Generation can be leveraged to enable fine-grained control on the different levels of difficulty and task similarity. 
\end{enumerate}
Specifically, in Section \ref{sec:generalization}, we briefly mentioned the inability of TRL to generalize to procedurally generated levels in 2D navigation~\cite{justesen2018illuminating}. Although it can generalize to different levels from one distribution, it can not handle arbitrary levels. 
% What are the trends we already see now and that are probably important in the near future?
One trend that could help overcome this problem, at least for navigation-related tasks, is the fusion of learning and planning, as in model-based reinforcement learning~\cite{sutton1991dyna}. Nevertheless, it is still an active research field with contributions such as a framework trying to unify the two~\cite{moerland2020framework}.
LSTMs also play an increasingly important role in improving generalization, as they can already enable the adaptability to different layouts of 2D navigation levels~\cite{duan2016rl2, sorokin2018episodic}. Furthermore, we view curriculum learning as a form of planning, as the creation of curricula inherently requires planning, and it has shown success as a TRL method~\cite{shao2018starcraft, green2019evolutionarilycurated}.

% mention trend towards benchmarks for TRL
Unfortunately, using benchmarks to compare novel approaches is not the norm in TRL yet. Contrary to Supervised Learning, where achieved accuracy percentages on well-known data sets can be perfectly compared, each sub-field and application would require different benchmarks to assess specific transfer capabilities of varying algorithms. But as we have shown (Section \ref{sec:analysis}), there are many experiments in similar fields like robotics or navigation that.
There are already interesting simulators like Mujoco~\cite{todorov2012mujoco} for continuous control, ProcGen~\cite{cobbe2019procgen} for generalization capabilities, Meta-World~\cite{yu2020meta} and Alchemy~\cite{wang2021alchemy} for meta-TRL.
However, there is still a plethora of other applications missing benchmarks to assess, e.g., game-playing AI~\cite{Volz2020TowardsGA} or 2D navigation. ProcGen does feature 2D Maze levels, but no benchmark verifies whether an algorithm can adapt to different movement styles (grid vs. continuous) or movement types (top-down vs. side-scroller).

Most experiments we encountered only transfer between fixed sets of tasks. As PCG has proven to be a reliable tool to improve generalization performance in RL~\cite{pcgrl, baker2020emergent}, the adoption of PCG is the logical next step for TRL. One could generate a seemingly infinite amount of different transfer tasks. 
Furthermore, PCG benchmarks would enable agents to control generation parameters that influence the difficulty, resulting in a dynamic curriculum that improves learning performance~\cite{green2019evolutionarilycurated}.
Moreover, the generation parameters could be chosen to influence task similarity. Such a quantifiable similarity control could be used as a benchmark metric to determine how much tasks may differ until the tested algorithm can not transfer efficiently anymore.

An ideal tool for the future would be a database, similar to OpenML~\cite{van2013openml}, that contains machine process-able information on all available TRL tasks/experiments. When approaching a new task, the trove of data could be used to cluster similar domains via task similarity metrics~\cite{carroll2005task}, to identify promising source tasks to transfer from. This would also allow a leader-board style comparison of how well which algorithm transfers between what tasks, like in GVGAI~\cite{perez2019general} per game (set).
While new task similarity metrics are still developed~\cite{ammar2014automated}, the problem identified by~\cite{carroll2005task} persists, that there is no one universal metric to encompass all similarity dimensions. The described database would enable the combination of all existing similarity metrics and the performance of different algorithms in transferring from one task to another to train a supervised network that outputs a singular numerical transfer success probability.

\section{Conclusion}
By borrowing methods and a dataset from the field of social network analysis (Section \ref{sec:method}), we have created a unique survey about Transfer in Reinforcement Learning (TRL). We collect tabular data about transfer-related metrics similar to~\cite{taylor-survey} but on a larger scale. We verified that out of 270 unique TRL entries, $\approx 74.8\%$ have not been included in any of the previous surveys~\cite{zhu-survey,lazaric-survey,taylor-intertask,bone-survey,silva-survey,taylor-survey,cur-survey,multitask-survey}. Because of the large scope, in which not every single entry can be mentioned textually, we created a website~\cite{interactivewebsite} that gives a better overview of the dataset with more graphs and the ability to interactively filter through the data.
With this data, we have underlined the large diversity of applications for TRL, which shows a focus on navigation, robotics, classic control, and games. 
We find that transfer, at least in RL, has a hard time generalizing to different problem variations (Section \ref{sec:generalization}). The transfer that works best is to problems that are similar. In tasks that involve planning, such as routes through a 2D levels, TRL lacks as it can not generalize to arbitrary layouts yet~\cite{justesen2018illuminating}. We do see an increase in methods that try to merge planning with learning (Section \ref{sec:outlook}) to overcome this limitation.
Another issue is the comparability of algorithms. Most approaches define their own slightly different version of known problems and compare their results to self-defined baselines. The field requires more benchmarks like Alchemy~\cite{wang2021alchemy}, Meta-World~\cite{yu2020meta}, or ProcGen~\cite{cobbe2019procgen} to quantify transfer performance properly and compare different algorithms. These benchmarks will increasingly include more procedural content generation to challenge generalization capabilities further. We provide a research agenda outlining how to achieve this goal.

\bibliographystyle{IEEETran}
\bibliography{report}
	
\end{document}